\documentclass[letterpaper, 10 pt, conference]{ieeeconf}  

\IEEEoverridecommandlockouts                              

\usepackage{amsmath} 
\usepackage{amssymb}  
\usepackage{cite}
\usepackage{amsfonts}
\usepackage{graphicx}
\usepackage{tikz}
\usepackage{booktabs}   
\usepackage{amsmath}    
\usepackage{multirow} 
\usepackage{algorithm}
\usepackage{algpseudocode}
\usepackage{kotex}
\usepackage{booktabs}
\usepackage{capt-of} 
\usepackage{adjustbox}
\usepackage{cuted}

\usepackage{wrapfig}
\usepackage{siunitx}
\usepackage{kotex}
\usepackage{url}

\title{\LARGE \bf
Collision Snapshot Guided Time-Reversed \\Safety-Critical Scenario Generation

}

\author{Taehyung Kim$^{1}$ and Jongeun Choi$^{1}$%
\thanks{$^{1}$Taehyung Kim and Jongeun Choi are with Yonsei University, Seoul, South Korea.
{\tt\small \{kth8606, jongeunchoi\}@yonsei.ac.kr}}%
}

\begin{document}
\maketitle
\thispagestyle{empty}
\pagestyle{empty}
\begin{abstract}
The generation of safety-critical traffic scenarios is essential for training and evaluating autonomous vehicles. Prior approaches typically perturb the trajectories of existing agents in a traffic scenario using simplified adversarial objectives to induce safety-critical interactions, which can limit the plausibility and diversity of the generated scenarios. Although inserting new adversarial vehicles can alleviate this limitation, determining when and where to introduce them in a scenario-specific manner remains challenging. In this work, we introduce \underline{CO}llision \underline{S}napshot guided \underline{T}im\underline{E}-\underline{R}eversed safety-critical scenario generation (COSTER), a framework that leverages learned traffic priors to determine plausible collision times and locations. COSTER first constructs a collision snapshot by inserting a new vehicle in contact with the target vehicle at the identified collision state within a traffic scenario. Starting from this collision snapshot, a conditional variational autoencoder is used to perform a time-reversed rollout, reconstructing the trajectory of the inserted vehicle backward toward earlier timesteps. Experiments show that COSTER outperforms existing methods in plausibility, diversity, and data efficiency. Moreover, agents trained on COSTER-generated scenarios reduce collision rates by 31\% on safety-critical scenarios from the Waymo Open Motion Dataset while also improving ego task completion. The project website is available at \url{https://anonym-121.github.io/COSTER/}.
\end{abstract}

\section{Introduction}
One of the major obstacles to progress toward fully autonomous driving is well illustrated by the concept of the \lq curse of rarity' proposed by Liu and Feng \cite{ref41}. The curse of rarity refers to the phenomenon where safety-critical events, resulting from complex combinations of diverse environmental factors, occur too infrequently in real-world scenarios. This scarcity poses a challenge for data-driven algorithms to model and generalize such events. 

One promising approach to addressing these challenges is to leverage simulation to generate large volumes of safety-critical scenarios for training \cite{ref18} and validation \cite{ref22}. Therefore, the generation of diverse and plausible safety-critical scenarios has been actively studied from multiple perspectives \cite{ref50, ref23}. Among the various approaches to safety-critical scenario generation, methods targeting the planning module of autonomous vehicles have primarily focused on introducing adversarial vehicles within the nominal traffic scenario to threaten the target vehicle.

A nominal traffic scenario can be transformed into a safety-critical scenario by selecting an existing vehicle and perturbing its trajectory so that it poses a threat to the target vehicle. However, this approach requires hand-crafted adversarial objectives, making it costly to design and limiting the diversity of collision scenarios. It also strongly depends on the original scenario configuration: unless the perturbed agent is already close enough to approach the target vehicle, the resulting behavior tends to be unrealistically aggressive or fails to produce a safety-critical scenario.

To address these limitations, recent approaches insert a new adversarial vehicle into the nominal scenario \cite{ref11}. To ensure a collision with the target vehicle, the adversarial vehicle is initialized at a selected time with a pose that collide with the target, and trajectory of adversarial vehicle is then generated backward in time. This reframing avoids the need to hand-craft an adversarial objective and removes the dependence on the base scene configuration, while guaranteeing a collision with the target vehicle. Despite its effectiveness, prior work determines the inserted vehicle using random pose offsets and a fixed collision time. Random pose placement may fail to account for the surrounding context, leading to overly aggressive scenarios or providing the time-reversed trajectory generation model with out-of-distribution initial conditions. Moreover, a fixed collision time may overlook more favorable opportunities for generating diverse and plausible collisions within the nominal scenario.

\begin{figure*}[!t]
    \centering
    \includegraphics[width=6.7in]{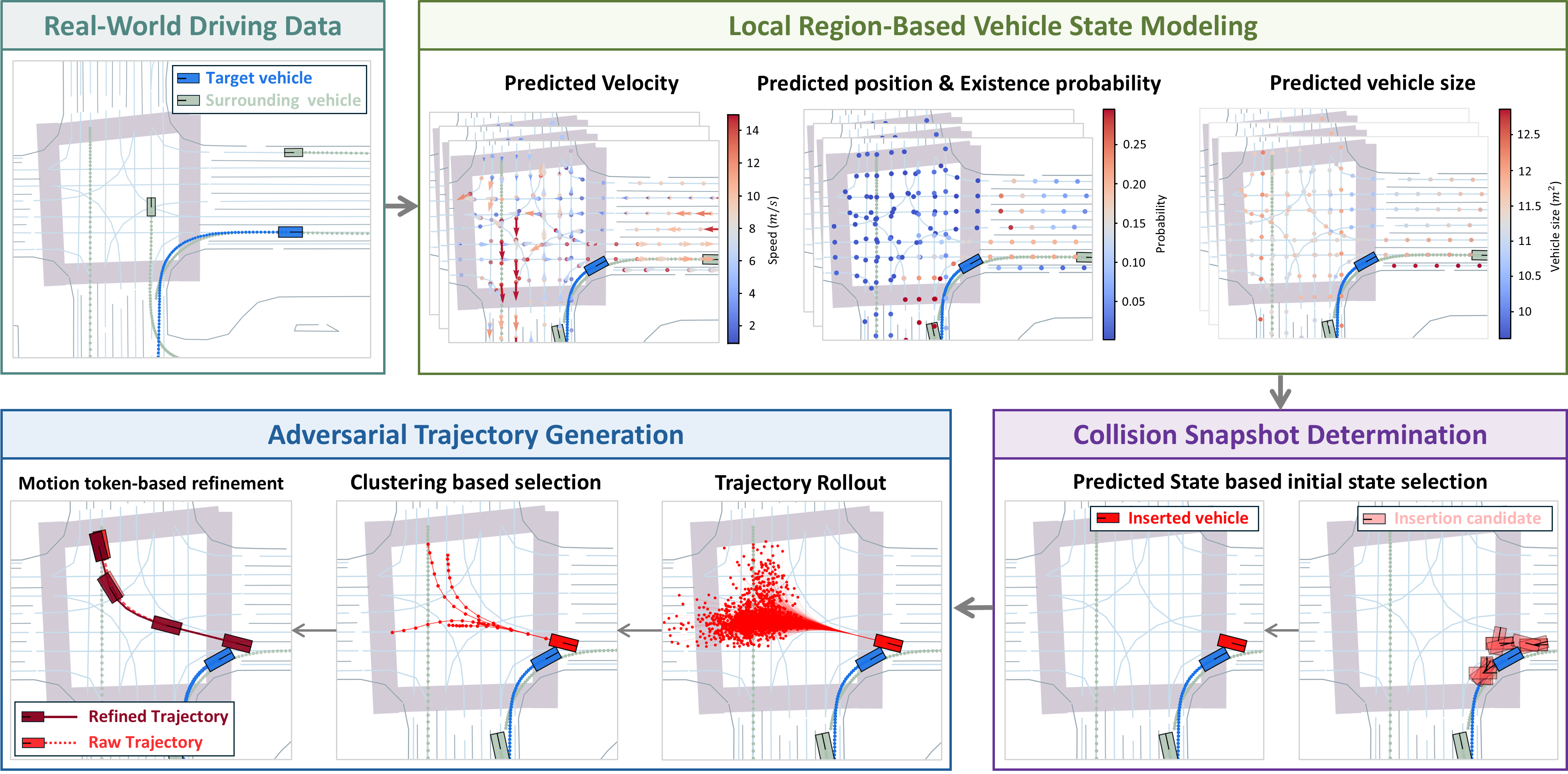}
    \caption{\textbf{The overview of COSTER.} COSTER predicts surrounding vehicle states at each timestep (Sec.~\ref{sec:2}), identifies the highest-risk timestep to construct a collision snapshot by inserting a contextually plausible adversarial vehicle in contact with the target vehicle (Sec.~\ref{sec:3}), and generates physically feasible safety-critical scenarios via time-reversed trajectory rollout with clustering-based selection and refinement (Sec.~\ref{sec:4}).}
    \label{fig_1}
\end{figure*}

To address this limitation, we introduce \textbf{CO}llision \textbf{S}napshot-guided \textbf{T}im\textbf{E}-\textbf{R}eversed scenario generation (COSTER). COSTER leverages direct access to collision snapshots in time-reversed generation, using traffic priors learned from the nominal traffic data to determine both the collision configuration and collision time. Compared with prior approaches that rely on random pose initialization and a fixed collision time, this traffic-prior-based design provides three key advantages: (i) it accounts for the surrounding traffic context to construct scenario-specific collision configurations; (ii) it reduces the likelihood of placing the inserted vehicle in an out-of-distribution state, enabling the subsequent generative module to reconstruct plausible trajectories from more realistic initial conditions; and (iii) it selects collision times according to the evolving traffic context, creating more diverse and plausible opportunities for safety-critical interactions. In addition, we propose a conditional variational autoencoder (CVAE) \cite{ref93} tailored for time-reversed trajectory generation, which updates surrounding-agent information at each timestep during the backward rollout from a collision snapshot. With these components, we aim to achieve diverse, and context-aware generation of time reversed trajectory. The overall framework is illustrated in Fig.~\ref{fig_1}.

Our main contributions can be summarized as follows:
\begin{itemize}
\item We propose COSTER, a collision-snapshot-guided time-reversed generation framework that directly incorporates scenario-specific traffic priors into collision configurations and time.

\item We introduce a velocity-entropy-based strategy for collision-time selection and a CVAE-based time-reversed rollout that updates surrounding-agent information at each timestep, enabling multimodal and context-aware adversarial trajectory generation.

\item We show that COSTER outperforms state-of-the-art methods in scenario quality and diversity, while agents trained on COSTER-generated scenarios achieve improved downstream driving safety.
\end{itemize}

\section{Related Work}

Early importance sampling \cite{ref21} and adaptive stress testing \cite{ref73} methods accelerate rare-event discovery but operate at the parameter level. Subsequent work shifted to trajectory-level perturbation, modifying recorded motions under kinematic constraints \cite{ref40} or differentiable model constraints \cite{ref5, ref38}. STRIVE \cite{ref1} encodes nominal scenarios into a CVAE latent space and optimizes adversarial objectives therein, while CAT \cite{ref4} decomposes the objective into motion-prediction sub-problems. However, these methods depend on simplified adversarial objectives (e.g., distance minimization \cite{ref1}, marginal collision probability \cite{ref4}) that do not account for scenario-specific context, resulting in limited diversity, overly aggressive and map-violating trajectories.

Several recent methods attempt to address these limitations. CaDRE \cite{ref6} uses quality-diversity optimization to broaden collision configurations, while GOOSE \cite{ref57} and SEAL \cite{ref7} improve behavioral realism via goal-conditioned and skill-based hierarchical RL, respectively. Despite these advances, each method retains notable limitations. CaDRE relies on a simplified distance-based objective with diversity bounded by its predefined measure space. GOOSE requires manual design and selection of an interaction mode (e.g., cut-in, deceleration), while SEAL depends on a pretrained CAT model. Both, along with CaDRE, modify existing agents, making their generation capacity heavily dependent on the initial configuration of the base scenario.

To address these limitations, Adv-BMT \cite{ref11} inserts a new adversarial vehicle into the given scenario. The vehicle is initialized at a future state that collides with the target vehicle, and its trajectory is then generated backward in time with bidirectional transformer. However, the adversarial vehicle is randomly initialized relative to the target vehicle, without explicitly accounting for the scenario-specific traffic context. Moreover, such random initialization can place the learned generative model under out-of-distribution initial conditions, often resulting in implausible trajectories and requiring extensive filtering. These limitations restrict the diversity and plausibility of the generated scenarios. In addition, Adv-BMT uses a fixed collision time rather than explicitly selecting a collision time suited to each scenario.

In contrast, COSTER leverages traffic priors to construct scenario-specific collision snapshots by estimating contextually plausible adversarial vehicle states in contact with the target vehicle. This allows the subsequent CVAE-based time-reversed rollout to start from an in-distribution initial state, avoiding map-inconsistent or physically infeasible initialization and enabling stable trajectory generation using a CVAE trained on nominal data. COSTER further selects the collision time for each scenario based on velocity entropy, aiming to improve data efficiency while promoting diverse and contextually plausible collision scenarios.

\section{Proposed Method}
\subsection{Problem Formulation}
A traffic scenario is denoted as $S_{1:T} = (M, \mathcal{T}_{1:T})$, where $M$ represents the vectorized map and $\mathcal{T}_{1:T} = \{ \tau_{1:T}^{1}, \ldots, \tau_{1:T}^{N} \}$ denotes the set of trajectories for $N$ vehicles over $T$ timesteps. Each trajectory $\tau_{1:T}^{n}$ consists of the vehicle states across time, represented as $\{ s_{1}^{n}, \ldots, s_{T}^{n} \}$. COSTER aims to generate a collision scenario $S' = \big(M,\; \mathcal{T}_{1:t_{\text{col}}}' \cup \tau_{1:t_{\text{col}}}^{\text{adv}} \big)$, where $t_{\text{col}}$ is the collision time and $\tau_{1:t_{\text{col}}}^{\text{adv}}$ is the trajectory of an inserted adversarial vehicle that induces safety-critical events with the target vehicle in the original scenario. COSTER constructs $S'$ in two stages. First, it selects a collision snapshot $\chi = (t_{\text{col}},\, s_{t_{\text{col}}}^{\text{adv}},\, s_{t_{\text{col}}}^{\text{tar}})$, where $s_{t_{\text{col}}}^{\text{adv}}$ and $s_{t_{\text{col}}}^{\text{tar}}$ denote the states of the adversarial and target vehicles at $t_{\text{col}}$. Second, a time-reversed trajectory distribution $p_\varphi$ is trained and used to sample trajectories of the adversarial vehicle $\tau_{1:t_{\text{col}}-1}^{\text{adv}}$ from $p_{\varphi}(\,\cdot \mid s_{t_{\text{col}}}^{\text{adv}},\, S_{1:t_{\text{col}}})$.

\subsection{Local Region-Based Vehicle State Modeling}
\label{sec:2}
\begin{figure}
    \centering
    \begin{tikzpicture}
        \node[anchor=south west, inner sep=0] (image) at (0,0)
            {\includegraphics[width=0.95\linewidth]{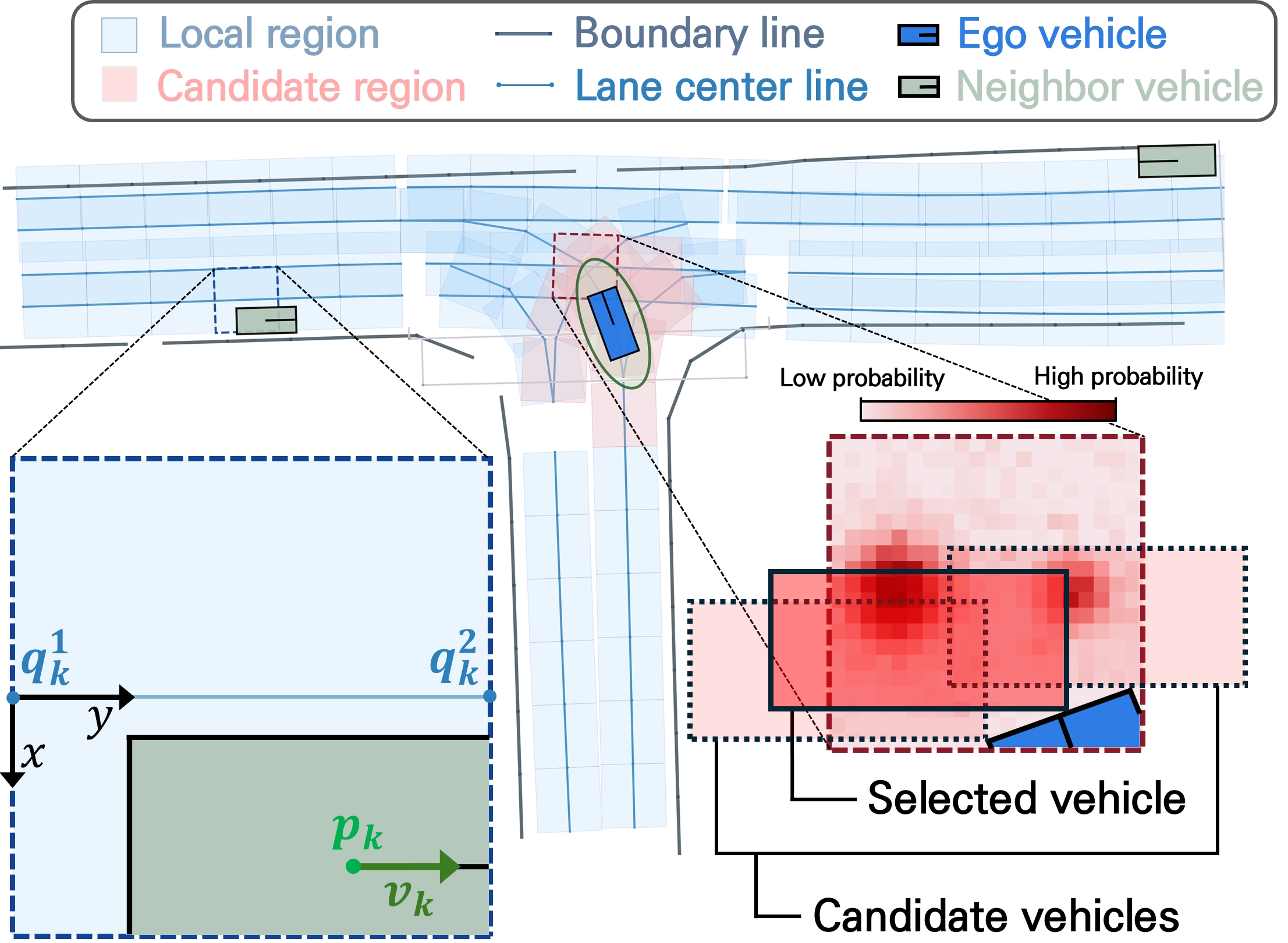}};
        \node at (1.6,-0.4) {\normalsize (a)};
        \node at (6.3,-0.4) {\normalsize (b)};
    \end{tikzpicture}
    \caption{Visualization of a sample scenario with overlaid local regions. (a) Region-wise representation. (b) Adversarial vehicle placement. At the given collision time, the peak-probability region within the elliptical area (red dashed line) is selected. The heading is then estimated at the red region, and the adversarial vehicle is translated into contact with the target vehicle to be placed at its final high-probability position.}
    \vspace{10pt}
    \label{fig2}
\end{figure}

To contextually initialize the adversarial vehicle, we partition \(M\) into local regions \(\{m_1, \ldots, m_K\}\) and estimate the expected vehicle state within each region. Large-scale real-world traffic datasets often provide vectorized $M$ with approximations of curved road elements, including lane boundaries, lane centerlines, and crosswalks. To partition the region, we uniformly sample points along the vectorized lane lines in the given \(M\), from which two adjacent points, \(q_k^1, q_k^2 \in \mathbb{R}^2\), are selected as shown in Fig.~\ref{fig2} (a). We define a local region coordinate by setting the origin at \(q_k^1\), aligning the \(y\)-axis with the direction of \(q_k^2 - q_k^1\), and constructing a rectangular region with a width of $w_r$ along the \(x\)-axis and a height of \(\lVert q_k^2 - q_k^1 \rVert\) along the \(y\)-axis. For each local region \(m_k\), the states of traffic agents passing through \(m_k\) are represented in the region's local coordinate system as a vector \(r_k\), defined as $r_k = (q_k^1, q_k^2, l_k, \sigma_k, p_k, v_k, b_k, \delta_k)$.

In this representation, $l_k$ denotes the lane type and $\sigma_k$ indicates the traffic light state. $p_k, v_k, b_k \in \mathbb{R}^2$ represent the position, velocity and size of the traffic agent, respectively. $\delta_k$ is an indicator variable that is 0 when no traffic agent is present in the region and 1 otherwise. The regional representations \(\{r_1, \ldots, r_K\}\) are first projected into a higher-dimensional feature space and then refined via multi-context gating (MCG)~\cite{ref19}, which approximates cross-attention~\cite{ref63} by conditioning on a single global context vector, offering significantly reduced computational cost while preserving expressive power~\cite{ref12, ref92}. The resulting context-aware features \(r_k'\) are decoded by independent MLP heads: one parameterizes a Bernoulli distribution over $\delta_k$ for placement prediction, while the others model $p_k$, $v_k$, and $b_k$ as discretized categorical distributions, from which continuous values are recovered via bin-center decoding at inference. Following prior works \cite{ref12,ref92}, we model each vehicle attribute with a separate distribution conditioned on shared contextual features, avoiding explicit modeling of the high-dimensional joint distribution. Let $y_{x,k}$ and $\hat{p}_{x,k}$ denote the discretized ground-truth bin index and the predicted categorical distribution for each state variable $x \in \{p, v, b\}$, corresponding to position, velocity, and size of region $k$, respectively. The model is trained end-to-end with the following loss:\begin{align}
\mathcal{L}
=\frac{1}{K}\sum_{k=1}^{K}
\left[
\mathrm{BCE}(\delta_k,\hat{\delta}_k)
+\delta_k
\sum_{x\in\{p,v,b\}}
\mathrm{CE}\!\left(
\mathrm{y}_{x, k},
\hat{\mathrm{p}}_{x, k}
\right)
\right].
\label{eq:loss_function_1}
\end{align}

\subsection{Collision Snapshot Determination}
\label{sec:3}
We now determine a collision snapshot $\chi = (t_{\text{col}}, s_{t_{\text{col}}}^{\text{adv}}, s_{t_{\text{col}}}^{\text{tar}})$ based on the estimated traffic agent states in the local regions. Unlike prior methods where collision time is implicitly determined as a byproduct of adversarial objectives, COSTER's snapshot-first design enables explicit selection of $t_{\text{col}}$ via any suitable risk criterion. In this work, we adopt velocity entropy $H$ as the selection criterion, motivated by its strong empirical correlation with collision risk~\cite{ref65, ref64, ref87, ref88}.

The regional traffic modeling described in Sec.~\ref{sec:2} is applied independently at each timestep in $S$, thereby characterizing a single scene. In this section, we extend the notation to incorporate time, and denote the velocity distribution of region $m_k$ at timestep $t$ as \(\hat{p}_{v,k,t}\). At each timestep, the set of local regions $\mathcal{N}_t^{\text{tar}}$ within an elliptical neighborhood centered at the target vehicle, with longitudinal and lateral axis lengths $\rho_{\text{long}}$ and $\rho_{\text{lat}}$, respectively, is selected as shown in Fig.~\ref{fig2} (b), and \(\hat{p}_{v,k,t}\) is estimated for each local region. We then compute \(H_t\) from the normalized state distribution $\bar{p}_{\text{v},t}$ obtained by uniformly averaging the distributions \(\hat{p}_{v,k,t}\) over the \(|\mathcal{N}_t^{\text{tar}}|\) regions. The velocity entropy at timestep $t$ is then computed as\[
H_t
=
- \sum_{i}
\bar{p}_{\text{v},t}(i)
\log
\bar{p}_{\text{v},t}(i),
\]where $i$ indexes the discretized velocity bins. The timestep with the largest \(H_t\) is selected as \(t_{\text{col}}\), and the state of the corresponding vehicle at that timestep is taken as \(s_{t_{\text{col}}}^{\text{tar}}\). 

Among the $|\mathcal{N}_{t_\text{col}}^{\text{tar}}|$ local regions around the target vehicle, we first identify the region with the highest $\mathrm{Bern}_{\text{pla},k,t_{\text{col}}}$, as described in Sec.~\ref{sec:2}. Within the selected region, the heading, velocity, and size of the adversarial vehicle are sampled from the predicted categorical distributions. The adversarial vehicle candidate is then translated toward the target vehicle while maintaining a fixed relative heading, until the two vehicles are in contact. Since multiple center positions may satisfy the contact condition, we select the final placement based on the maximum-probability position under the predicted categorical position distribution. This determines $s_{t_{\text{col}}}^{\text{adv}}$, completing $\chi$.

\subsection{Adversarial Trajectory Generation}
\label{sec:4}
Based on $\chi$, we reconstruct the adversarial vehicle trajectory prior to $t_{\text{col}}$ using a model trained on a large-scale traffic dataset to recover vehicle trajectories backward from a generic reverse endpoint $t_{\text{rev}}$ to $t_1$, with $t_{\text{rev}}=t_\text{col}$ at generation time. Motivated by the success of CVAEs in generating multimodal trajectories~\cite{ref69, ref72}, we adopt a CVAE-based framework for COSTER. The CVAE in COSTER takes $\chi$, $M$, $\mathcal{T}$ and a latent variable as input, all expressed in coordinates relative to the vehicle being reconstructed. $M$ is encoded into map embeddings, which serve as keys and values in cross-attention queried by the reconstructed vehicle state. The resulting representation is concatenated with the vehicle-state embedding to form the initial decoder input $h_{t_\text{rev}}$.

However, the context at \( t_\text{rev} \) alone is insufficient for accurate prediction of the subsequent trajectory. We therefore propose dynamic surrounding attention (DSA) in the CVAE decoder to update the surrounding-agent context at every timestep. Specifically, the model predicts the reverse displacement at timestep $t$ using the decoder input representation $h_{t+1}$. Based on the predicted state, the surrounding-agent context is dynamically updated and incorporated into $h_{t+1}$ through an RNN. The surrounding-agent context is computed via a cross-attention module between the vehicle being reconstructed and its neighboring agents, where $h_{t+1}$ serves as the query, and the relative position and velocity differences between the vehicle and its neighbors serve as both keys and values. The resulting attention output is concatenated with the state of the vehicle being reconstructed at the corresponding time step to form the surrounding-agent context. Despite the limited information in $\chi$, DSA enables efficient generation of multimodal, context-aware adversarial-vehicle trajectories, while retaining the computational advantage of sampling-based generation. Refer to Table~\ref{tab:1} for the ablation and performance results of DSA.

During training, reconstructs trajectories from $t_{rev}$ to $t_1$. At each decoding step, a latent variable $z_t$ is sampled from the posterior distribution, and the decoder, conditioned on both the decoder input representation $h_{t+1}$ and $z_t$, is trained to predict the displacement $d_t$. For posterior estimation, we additionally use $g_t$, an RNN-encoded summary of the ground-truth trajectory segment available only during training, enabling the model to better capture long-horizon strategic behaviors. The conditional prior \( p_\theta \), posterior \( q_\phi \), and likelihood \( p_\xi \) are jointly optimized by maximizing the following evidence lower bound (ELBO), averaged over $L=t_{rev}-t_1$: \begin{align}
\max_{\phi,\theta,\xi}\; \mathcal{L}
={}&\mathbb{E}_i\Biggl[\frac{1}{L}\sum_{t=t_1}^{t_{\text{rev}}-1}
     \mathbb{E}_{q_\phi}\Bigl[
     \underbrace{\log p_\xi(d_t\mid z_t,h_{t+1})}_{\text{1st term}} \nonumber\\[-2pt]
   &\hspace{4em}
     -\underbrace{\log\tfrac{q_\phi(z_t\mid g_t,h_{t+1})}
                            {p_\theta(z_t\mid h_{t+1})}}_{\text{2nd term}}
     \Bigr]\Biggr].
\label{eq:rec}
\end{align} The first term of Eq.~(\ref{eq:rec}) maximizes the reconstruction log-likelihood, while the second term regularizes the posterior toward the prior via KL divergence. During inference, the decoder utilizes $z_t$ sampled from the prior distribution. 

The trajectories of adversarial vehicle are then oversampled from the CVAE and refined through two stages: (i) K-means clustering on lateral-displacement and arc-length features for representative selection, efficiently summarizing multimodal behavior while reducing sensitivity to outliers~\cite{ref72}, and (ii) projecting each state transition of the selected representatives onto its nearest motion token in a motion vocabulary constructed following~\cite{ref3} to ensure physical feasibility.

\section{Experiments}
\subsection{Experiment Setting}
We train the models described in Sec.~\ref{sec:2} and Sec.~\ref{sec:4} using the Waymo Open Motion Dataset v1.1 (WOMD), which contains 9-second scenarios (530,000 scenes). The local region width is set to $w_r = 5\,\text{m}$. The longitudinal and lateral semi-axis lengths, $\rho_{\mathrm{long}}$ and $\rho_{\mathrm{lat}}$, are set to twice the ego vehicle’s length and width, respectively. The trajectory generation horizon in Sec.~\ref{sec:4} is adaptively determined based on $t_{\mathrm{col}}$. The original dataset, recorded at 10 Hz, is downsampled to 2 Hz, and the motion vocabulary is constructed accordingly.

\begin{table*}[t!]
\centering
\small
\caption{Quantitative Comparison on Past Trajectory Prediction Accuracy}
\label{tab:1}
\begin{tabular}{lccccc}
\toprule
 & \textbf{minADE (m)} $\downarrow$ 
 & \textbf{minFDE (m)} $\downarrow$ 
 & \textbf{avgADE (m)} $\downarrow$
 & \textbf{avgFDE (m)} $\downarrow$
 & \textbf{Inference Time (ms)} $\downarrow$ \\
\midrule
\noalign{\global\aboverulesep=0pt \global\belowrulesep=0pt}%
Constant Velocity        & 1.695 & 4.961 & 1.695 & 4.961 & - \\[-1.5pt]
Constant Lane            & 14.860 & 17.442 & 14.860 & 17.442 & - \\[-1.5pt]
BMT \cite{ref11}    & 0.417 & 1.022 & 0.838 & 2.190 & 536.2 \\[-1.5pt]
COSTER (w/o DSA)         & 0.557 & 1.545 & 1.212 & 3.460 & \textbf{5.7} \\[-1.5pt]
\textbf{COSTER (w/ DSA)} & \textbf{0.259} & \textbf{0.828} & \textbf{0.570} & \textbf{1.966} & 39.4 \\[-1.5pt]
\bottomrule
\end{tabular}
\global\aboverulesep=0.4ex \global\belowrulesep=0.65ex
\end{table*}

\begin{table}[ht]
\centering
\small
\setlength{\tabcolsep}{4pt}
\caption{MMD Comparison on Vehicle State Inference.}
\begin{tabular}{lcccc}
\toprule
 & \textbf{Position} $\downarrow$ & \textbf{Heading} $\downarrow$ & \textbf{Speed} $\downarrow$ & \textbf{Size} $\downarrow$ \\
\midrule
\noalign{\global\aboverulesep=0pt \global\belowrulesep=0pt}%
Mean Value & 0.2249 & 0.4237 & 0.1564 & 0.5451 \\[-1.5pt]
LCTGen \cite{ref92} & 0.1466 & 0.1555 & 0.2286 & 0.1052 \\[-1.5pt]
TrafficGen \cite{ref12} & 0.1367 & 0.1277 & 0.1704 & 0.0937 \\[-1.5pt]
\textbf{COSTER} & \textbf{0.1242} & \textbf{0.1116} & \textbf{0.1399} & \textbf{0.0764} \\[-1.5pt]
\bottomrule
\end{tabular}
\label{tab:mmd}
\end{table}

\subsection{Vehicle State Prediction Assessment}
\label{traffic_prior}
We evaluate the implementation described in Sec.~\ref{sec:2} against other baselines. For each method, we performed state inference on an identical set of 1,000 WOMD scenarios and measured the respective accuracy.

\noindent \textbf{Baseline.} Since the component implemented in Sec.~\ref{sec:2} has primarily been used as a submodule in prior frameworks, we adopt the corresponding modules from existing works as baselines. \textbf{TrafficGen} \cite{ref12} predicts regional vehicle states using an MLP-based Gaussian mixture model (GMM) over position, heading, speed, and size. \textbf{LCTGen} \cite{ref92} similarly adopts GMM-based modeling but incorporates interaction-aware agent features through self- and cross-attention. For a fair comparison, we remove the language conditioning from the original LCTGen and rely only on map and agent features. COSTER represents each state variable as a discretized categorical distribution over bins. We additionally include a \textbf{Mean Value} baseline that predicts the average vehicle states computed from the same WOMD scenarios.

\noindent \textbf{Evaluation metrics.} We evaluate the predicted vehicle-state distributions using maximum mean discrepancy (MMD) \cite{ref104}, following prior traffic regional modeling studies \cite{ref12, ref92}. Lower MMD indicates a closer match to the ground-truth distribution.

\noindent \textbf{Analysis.} As shown in Table~\ref{tab:mmd}, the bin-based approach achieves consistently lower MMD than the GMM-based baselines across all state attributes. This suggests that discretized categorical modeling better captures complex vehicle-state distributions without imposing Gaussian assumptions or a fixed number of mixture components. In contrast, both TrafficGen and LCTGen are limited to five Gaussian components, which may limit their ability to represent irregular or highly multimodal distributions. Moreover, the discrete formulation enables the velocity entropy in Sec.~\ref{sec:3} to be computed in closed form, avoiding the numerical approximation required for GMM entropy estimation.

\subsection{Time-Reversed Trajectory Generation Assessment}
\label{sec:4_2}
We evaluate the time-reversed trajectory generation module by reconstructing past 5-second trajectories of existing vehicles in the WOMD.

\noindent \textbf{Baseline}. We compare COSTER without DSA (\textbf{COSTER (w/o DSA)}) and with DSA (\textbf{COSTER (w/ DSA)}) against the following baselines: \textbf{Constant Velocity}: linearly extrapolates the vehicle state at $t_\text{rev}$ backward assuming constant velocity. \textbf{Constant Lane}: extrapolates the vehicle backward along the nearest lane centerline at constant speed. \textbf{BMT}~\cite{ref11}: a bidirectional motion transformer for trajectory reconstruction.

\noindent \textbf{Evaluation metrics}. We report (\romannumeral 1) minADE, (\romannumeral 2) minFDE~\cite{ref101}, the minimum average and final displacement errors over 5 samples, (\romannumeral 3) avgADE, (\romannumeral 4) avgFDE~\cite{ref102}, the mean average and final displacement errors, and (\romannumeral 5) Inference Time per trajectory.

\noindent \textbf{Analysis}. As shown in Table~\ref{tab:1}, COSTER (w/ DSA) achieves the lowest displacement errors across all metrics, confirming its advantage over existing methods in both overall trajectory accuracy and final position prediction. Ablating DSA  confirms that dynamically updating surrounding-agent context at each decoding step is essential, particularly for reducing long-horizon drift as reflected in minFDE and avgFDE. Both COSTER variants also achieve significantly faster inference than BMT.

\subsection{Safety-Critical Scenario Assessment}
\label{sec:4_3}
We evaluate the full scenario generation pipeline by inserting adversarial vehicles into WOMD.

\begin{table*}[!t]
\centering
\setlength{\tabcolsep}{3pt}
\small
\caption{Quantitative Comparison of Generated Safety-Critical Scenarios.}
\label{tab:cat_cupids}
\begin{tabular}{lccccccc}
\toprule
 & \multicolumn{1}{c}{\textbf{Data Efficiency}} 
 & \multicolumn{1}{c}{\textbf{Adversariality}} 
 & \multicolumn{3}{c}{\textbf{Plausibility}} 
 & \multicolumn{2}{c}{\textbf{Diversity}} \\ 
\cmidrule(lr){2-2} \cmidrule(lr){3-3} \cmidrule(lr){4-6} \cmidrule(lr){7-8}
\textbf{} & \textbf{SUR (\%) $\uparrow$} 
& \textbf{CR (\%) $\uparrow$} 
& \textbf{ORR (\%) $\downarrow$} 
& \textbf{VDD $\downarrow$} 
& \textbf{ADD $\downarrow$}  
& \textbf{CSD (m/s) $\uparrow$} 
& \textbf{CPD (\si{\degree}) $\uparrow$} \\
\midrule
\noalign{\global\aboverulesep=0pt \global\belowrulesep=0pt}%
CAT \cite{ref4}    & 8.9 & 53.2 & 12.3 & 0.2775 & 0.2370 & 3.14 & 80.1 \\[-1.3pt]
SEAL \cite{ref7}   & 8.7 & 60.8 & 19.1 & 0.2828 & 0.2419 & 3.31 & 70.8 \\[-1.3pt]
Adv-BMT \cite{ref11} & 61.2 & \textbf{100} & 62.3 & \textbf{0.2394} & 0.3019 & 3.37 & 29.9 \\[-1.3pt]
\textbf{COSTER (Ours)} & \textbf{74.1} & \textbf{100} & \textbf{10.2} & 0.2485 & \textbf{0.0983} & \textbf{4.75} & \textbf{86.2} \\[-1.3pt]
\cmidrule(lr){1-8}
Random + CVAE & 63.5 & 100 & 51.8 & 0.2451 & 0.1147 & 3.42 & 31.6 \\[-1.3pt]
Snapshot + BMT & 71.8 & 100 & 19.4 & 0.2437 & 0.2864 & 4.68 & 84.1 \\[-1.3pt]
\bottomrule
\end{tabular}
\global\aboverulesep=0.4ex \global\belowrulesep=0.65ex
\end{table*}

\begin{figure*}[t!]
  \centering
  \begin{tikzpicture}
    \node[anchor=south west, inner sep=0] (image) at (0,0)
      {\includegraphics[width=5.5in]{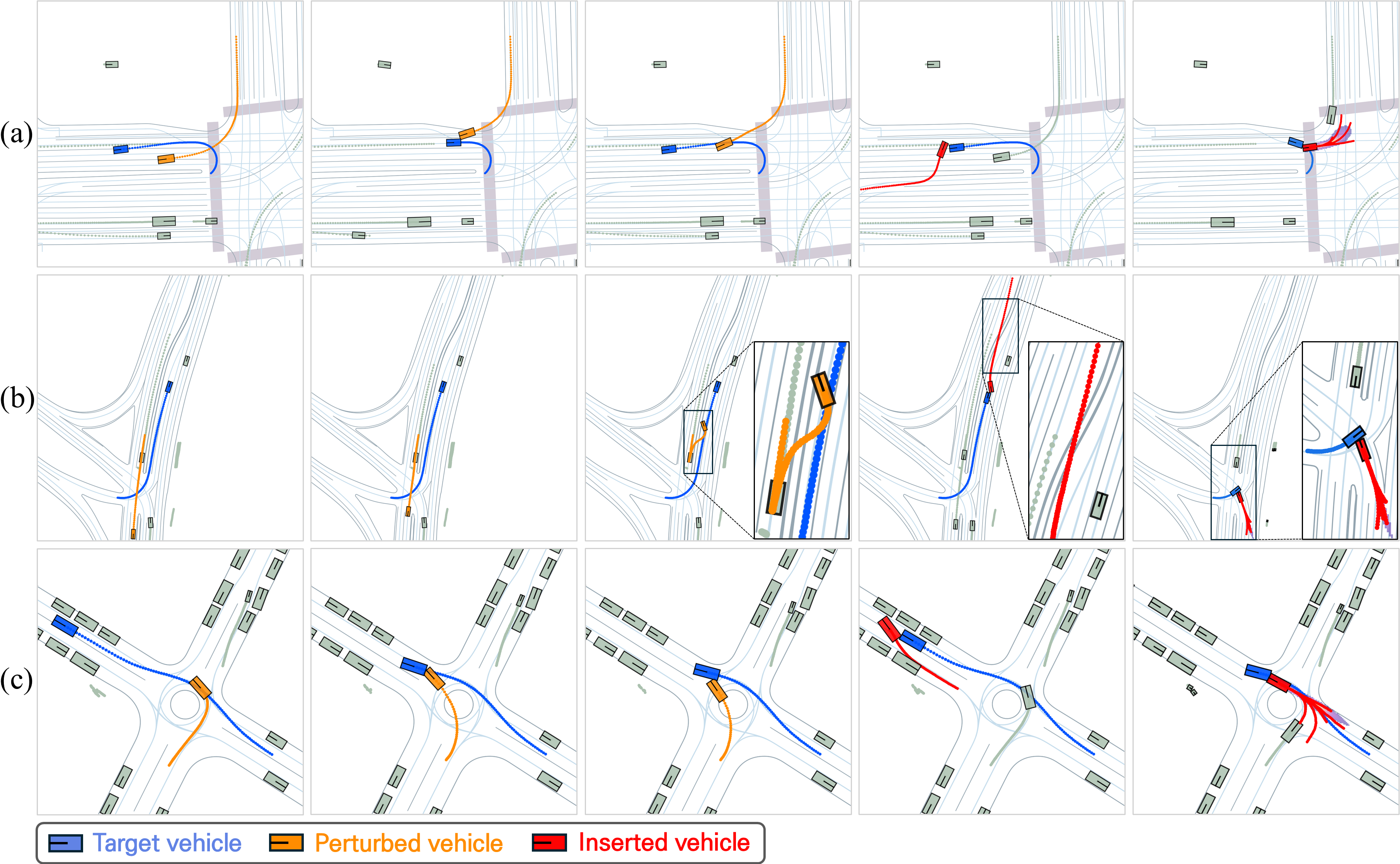}};

    \begin{scope}[x={(image.south east)}, y={(image.north west)}]

      \node[anchor=north, font=\small]
        at ({0.12}, {1+0.06}) {Original Scenario};

      \node[anchor=north, font=\small]
        at ({0.315}, {1+0.06}) {CAT \cite{ref4}};

      \node[anchor=north, font=\small]
        at ({0.518}, {1+0.06}) {SEAL \cite{ref7}};

      \node[anchor=north, font=\small]
        at ({0.71}, {1+0.06}) {Adv-BMT \cite{ref11}};

      \node[anchor=north, font=\small]
        at ({0.895}, {1+0.06}) {\textbf{COSTER}};

    \end{scope}
  \end{tikzpicture}
  \caption{Qualitative comparison of generated safety-critical scenarios.}
  \label{fig_3}
\end{figure*}
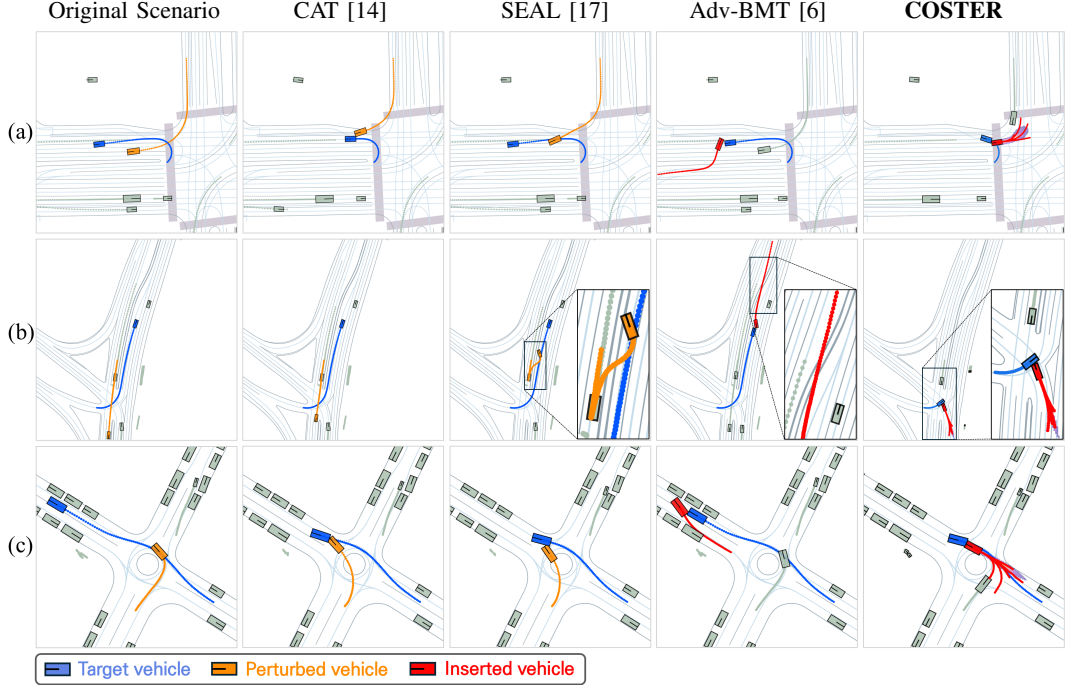

\noindent \textbf{Baseline}. We conduct both quantitative and qualitative comparisons with the following baselines: \textbf{CAT}~\cite{ref4}: Perturbs existing traffic agents by selecting collision-prone trajectories to generate collision scenarios based on a simplified collision objective. \textbf{SEAL}~\cite{ref7}: Perturbs existing traffic agents using skill-based policies to generate collision. \textbf{Adv-BMT}~\cite{ref11}: Inserts new adversarial vehicles and employs a bidirectional motion transformer to reconstruct trajectories. For a fair comparison, we fix the target vehicle as the ego vehicle across all methods. 

\noindent \textbf{Additional variants}. COSTER and Adv-BMT differ in two components: the collision snapshot and the reverse rollout model. To disentangle them, we evaluate the two hybrid configurations. \textbf{Random + CVAE} replaces our snapshot construction with the ego-pose-offset initialization of Adv-BMT; \textbf{Snapshot + BMT} feeds our snapshot to the BMT with its native post-processing.

\begin{table}[t!]
\centering
\small
\caption{Comparison of Collision Time Determination.}
\label{tab:abl}
\begin{tabular}{lcc}
\toprule
 & \textbf{CSD (m/s) $\uparrow$} & \textbf{CPD (\si{\degree}) $\uparrow$} \\
\midrule
Random timestep      & 4.12 & 70.2 \\
Last timestep       & 4.29 & 73.3 \\
\textbf{Velocity entropy timestep}   & \textbf{4.75} &  \textbf{86.2} \\
\bottomrule
\end{tabular}
\end{table}

\begin{table*}[t!]
\centering
\small
\caption{Downstream Evaluation of Policy Safety under Normal and Hard Scenarios}
\label{tab:womd_real}
\begin{tabular}{lcccccc}
\toprule
& \multicolumn{3}{c}{\textbf{WOMD-Normal}} & \multicolumn{3}{c}{\textbf{WOMD-Hard~\cite{ref95}}} \\
\cmidrule(lr){2-4} \cmidrule(lr){5-7}
\textbf{Training Data}
& \textbf{RCR $\uparrow$} & \textbf{CR $\downarrow$} & \textbf{ORR $\downarrow$}
& \textbf{RCR $\uparrow$} & \textbf{CR $\downarrow$} & \textbf{ORR $\downarrow$} \\
\midrule
\noalign{\global\aboverulesep=0pt \global\belowrulesep=0pt}%
WOMD-Only        & 0.68{\scriptsize\textpm0.03} & 0.17{\scriptsize\textpm0.01} & 0.24{\scriptsize\textpm0.02} & 0.30{\scriptsize\textpm0.02} & 0.32{\scriptsize\textpm0.03} & 0.32{\scriptsize\textpm0.04} \\[-1.5pt]
CAT \cite{ref4}          & 0.67{\scriptsize\textpm0.05} & 0.18{\scriptsize\textpm0.01} & 0.26{\scriptsize\textpm0.02} & 0.29{\scriptsize\textpm0.04} & 0.33{\scriptsize\textpm0.03} & 0.34{\scriptsize\textpm0.01} \\[-1.5pt]
SEAL \cite{ref7}         & 0.69{\scriptsize\textpm0.03} & 0.16{\scriptsize\textpm0.01} & 0.26{\scriptsize\textpm0.02} & 0.35{\scriptsize\textpm0.01} & 0.27{\scriptsize\textpm0.02} & \textbf{0.30}{\scriptsize\textpm0.03} \\[-1.5pt]
Adv-BMT \cite{ref11}        & 0.64{\scriptsize\textpm0.07} & 0.20{\scriptsize\textpm0.02} & 0.29{\scriptsize\textpm0.02} & 0.26{\scriptsize\textpm0.02} & 0.38{\scriptsize\textpm0.04} & 0.36{\scriptsize\textpm0.01} \\[-1.5pt]
\textbf{COSTER (w/o Aug)}        & 0.71{\scriptsize\textpm0.06} & \textbf{0.15}{\scriptsize\textpm0.03} & 0.23{\scriptsize\textpm0.04} & 0.38{\scriptsize\textpm0.04} & 0.24{\scriptsize\textpm0.01} & 0.31{\scriptsize\textpm0.02} \\[-1.5pt]
\textbf{COSTER (w/ Aug)}        & \textbf{0.73}{\scriptsize\textpm0.04} & \textbf{0.15}{\scriptsize\textpm0.01} & \textbf{0.22}{\scriptsize\textpm0.02} & \textbf{0.41}{\scriptsize\textpm0.02} & \textbf{0.22}{\scriptsize\textpm0.02} & \textbf{0.30}{\scriptsize\textpm0.03} \\[-1.5pt]
\bottomrule
\end{tabular}
\global\aboverulesep=0.4ex \global\belowrulesep=0.65ex
\end{table*}

\noindent \textbf{Evaluation metrics}. We adopt seven evaluation metrics based on prior studies \cite{ref1, ref9}. (\romannumeral 1) Scenario utilization rate (SUR), which measures the proportion of raw scenarios from which a collision scenario can be generated. (\romannumeral 2) Collision rate (CR), which measures the probability of a collision between the target and the adversarial vehicle. (\romannumeral 3) Collision point diversity (CPD), which measures the standard deviation of normalized collision points in the ego-centric coordinate frame. (\romannumeral 4) Collision speed diversity (CSD), which measures the standard deviation of relative collision speeds at the moment of impact across scenarios. (\romannumeral 5) Off road rate (ORR), which measures the proportion of trajectories in which the adversarial vehicle crosses the road boundary. (\romannumeral 6) Velocity distribution divergence (VDD) and (\romannumeral 7) acceleration distribution divergence (ADD), which measure the Jensen–Shannon divergence between the velocity and acceleration distributions of trajectories from raw scenarios and generated trajectories, respectively.

\noindent \textbf{Quantitative analysis}. As shown in Table~\ref{tab:cat_cupids}, COSTER achieves the highest SUR while matching the best CR. CAT and SEAL show low SUR due to their reliance on perturbing existing vehicles, limiting generation to scenarios that satisfy specific conditions. While Adv-BMT attains comparable SUR and CR, it relies on randomly initialized collision snapshots. As a result, the trajectory generation module trained on nominal data is frequently initialized in implausible or out-of-distribution states, leading to degraded realism, as evidenced by its high ORR. Similar limitations are observed in diversity. CAT and SEAL are inherently constrained by raw scenario configuration, while Adv-BMT uses random initialization constructed via ego-pose offsets, which yields limited variations in collision configurations. In contrast, COSTER achieves the highest CPD and CSD, indicating substantially richer diversity in both collision location and dynamics, while maintaining strong plausibility with the lowest ORR and ADD. These results suggest that constructing probabilistically grounded collision snapshots, followed by generative rollout, provides an effective approach for synthesizing safety-critical scenarios as out-of-distribution events from nominal data.

The hybrid variants further clarify the respective roles of collision initialization and trajectory rollout. Although BMT demonstrates strong trajectory reconstruction capability under nominal conditions, as shown in Table~\ref{tab:1}, Random + CVAE still exhibits substantially degraded plausibility, particularly in terms of ORR. This suggests that random collision initialization can expose the rollout model to implausible or out-of-distribution initial conditions, leading to unrealistic trajectories. In contrast, Snapshot + BMT markedly alleviates these failures, highlighting the importance of context-aware collision initialization, while the full COSTER further improves trajectory plausibility through its CVAE-based rollout. This failure mode is also qualitatively evident in Fig.~\ref{fig_3}, where the randomly initialized Adv-BMT produces implausible trajectories.

\noindent \textbf{Qualitative analysis}. Figure~\ref{fig_3} compares safety-critical scenarios generated from the same original scenario. In (a) and (c), CAT and SEAL produce perturbed  trajectories that leave the map, indicating that the adversarial objective based method often fails to account for diverse and context-aware situations. In (b), CAT fails to induce a collision under the given raw scenario configuration, while SEAL generates unrealistic behavior by failing to respond properly to target vehicle motion. Adv-BMT also shows some limitations. In (a) and (b), implausible collision snapshots from random initialization cause the generative model to fail, as discussed in the quantitative analysis section. In (c), it produces scenarios that do not adequately consider surrounding vehicles. In contrast, COSTER generates multimodal and plausible collision scenarios that account for interactions with surrounding agents. Beyond the five representative trajectories, the distribution of remaining generated trajectories is visualized as a heatmap using Gaussian kernel density estimation.

\subsection{Analysis on Velocity Entropy-based Selection}
We conduct an additional ablation study to more precisely isolate the impact of the collision time selection method.

\noindent \textbf{Baseline}. We compare three collision time determination strategies: \textbf{Random timestep}, which randomly samples a timestep from the base scenario, \textbf{Last timestep}, which fixes the collision time to the last timestep, and \textbf{Velocity entropy timestep}, which selects the timestep with the highest velocity entropy. For each strategy, safety-critical scenarios are generated using the COSTER pipeline from the same set of 1,000 WOMD scenarios.

\noindent \textbf{Evaluation metrics}. Among the metrics used in Sec.~\ref{sec:4_3}, we focus on CPD and CSD, as these are most directly influenced by the choice of collision time. Metrics that are largely insensitive to collision time determination (e.g., SUR, CR) are not reported for this experiment.

\noindent \textbf{Quantitative analysis}. As shown in Table~\ref{tab:abl}, both the random timestep and last timestep strategies yield lower CPD and CSD compared to velocity entropy-based selection, indicating more limited collision configuration diversity. This result suggests that velocity entropy-based timestep selection more effectively identifies regions of complex traffic interaction, where a wider variety of candidate adversarial vehicle states are available. The richer candidate pool in such regions naturally translates into a broader range of collision types in terms of both impact location and relative speed.

\subsection{Downstream Evaluation on Policy Safety}
\label{sec:4_4}
We evaluate whether scenarios generated by COSTER lead to greater improvements in downstream task performance compared to those generated by CAT, SEAL, and Adv-BMT. To this end, RL-based traffic agents are trained in closed-loop simulation using MetaDrive~\cite{ref96} with 1,000 scenarios produced by each method. We follow the RL training configuration of Liu et al.~\cite{ref11}. COSTER generates multi-modal trajectories from a single collision snapshot. We consider two variants: COSTER (w/ Aug), where 1000 trajectories are oversampled and five clustered trajectories are used for training and COSTER (w/o Aug), where a single representative trajectory is selected from the five trajectories. Additionally, we introduce a control baseline, \textbf{WOMD-Only}, where the agent is trained on 1000 unmodified WOMD scenarios without any adversarial perturbations.

We evaluate under two settings. (\romannumeral 1) \textbf{WOMD-Normal:} A generalization setting with 200 randomly sampled scenarios from the WOMD not used in training, to assess potential performance degradation in normal conditions. (\romannumeral 2) \textbf{WOMD-Hard:} A safety-critical setting consisting of scenarios selected using SafeShift \cite{ref95} hierarchical scoring (top 20 percentile) from the WOMD, as used in previous work~\cite{ref7}, to measure performance in high-risk environments. As evaluation metrics, we use route completion rate (RCR), defined as the proportion of scenarios in which the trained agent successfully completes the route without collision or off-road deviation, along with CR and ORR.

\textbf{Analysis}. As shown in Table~\ref{tab:womd_real}, COSTER consistently outperforms prior methods across both evaluation sets. On WOMD-Normal, COSTER achieves the highest RCR while maintaining low CR and ORR, indicating that it preserves performance in standard scenarios without degradation. More importantly, on the more challenging WOMD-Hard split, COSTER substantially improves RCR, outperforming baseline methods while also achieving the lowest CR. Notably, COSTER (w/ Aug) demonstrates superior performance compared to COSTER (w/o Aug) in SafeShift scenarios. Overall, these results suggest that the scenarios generated by COSTER not only contribute effectively to training compared to baseline methods, but also facilitate more efficient learning by providing diverse variants of safety-critical scenarios derived from a single raw scenario.  

\section{Conclusion}
We presented COSTER, a safety-critical scenario generation framework that constructs collision time and configuration from learned traffic priors and generates adversarial trajectories via time-reversed CVAE rollout. By controlling collision configurations, COSTER improves plausibility, diversity, data efficiency, and downstream driving safety. Despite these benefits, COSTER is currently limited to vehicle-to-vehicle collisions and may overlook hazardous low-entropy scenarios. Future work will extend the framework to diverse road users and incorporate richer semantic priors, such as vision–language models.

\addtolength{\textheight}{-0cm}

\bibliographystyle{IEEEtran}
\bibliography{references}

\end{document}